\documentclass[10pt,twocolumn,letterpaper]{article}

\usepackage{iccv}
\usepackage{times}
\usepackage{epsfig}
\usepackage{graphicx}
\usepackage{amsmath}
\usepackage{amssymb}

\usepackage{booktabs}
\usepackage[dvipsnames]{xcolor}
\newcommand{\dho}[1]{\ifdefined\comments {\color{PineGreen} #1} \else #1 \fi}

\usepackage[pagebackref=true,breaklinks=true,letterpaper=true,colorlinks,bookmarks=false]{hyperref}

\iccvfinalcopy 

\def\iccvPaperID{10563} 

\ificcvfinal\fi

\begin{document}

\title{Leveraging Spatial and Photometric Context for Calibrated Non-Lambertian Photometric Stereo}

\author{David Honz\'{a}tko\\
CSEM\\
Neuch\^{a}tel, Switzerland\\
{\tt\small david.honzatko@csem.ch}
\and
Engin~T\"{u}retken\\
CSEM\\
Neuch\^{a}tel, Switzerland\\
{\tt\small engin.tueretken@csem.ch}
\and
Pascal Fua\\
EPFL\\
Lausanne, Switzerland\\
{\tt\small pascal.fua@epfl.ch}
\and
L. Andrea Dunbar\\
CSEM\\
Neuch\^{a}tel, Switzerland\\
{\tt\small andrea.dunbar@csem.ch}
}

\maketitle
\ificcvfinal\thispagestyle{empty}\fi

\begin{abstract}
    The problem of estimating a surface shape from its observed reflectance properties still remains a challenging task in computer vision. The presence of global illumination effects such as inter-reflections or cast shadows makes the task particularly difficult for non-convex real-world surfaces.
    State-of-the-art methods for calibrated photometric stereo address these issues using convolutional neural networks (CNNs) that primarily aim to capture either the spatial context among adjacent pixels or the photometric one formed by illuminating a sample from adjacent directions.
   
    In this paper, we bridge these two objectives and introduce an efficient fully-convolutional architecture that can leverage both spatial and photometric context simultaneously. In contrast to existing approaches that rely on standard 2D CNNs and regress directly to surface normals, we argue that using separable 4D convolutions and regressing to 2D Gaussian heat-maps severely reduces the size of the network and makes inference more efficient. 
    Our experimental results on a real-world photometric stereo benchmark show that the proposed approach outperforms the existing methods both in efficiency and accuracy.
    
\end{abstract}

\section{Introduction}
\label{sec:introduction}

The general problem of calibrated photometric stereo involves estimating an object shape given an a priori unknown number of images of the object captured from the same viewpoint but illuminated from a different angle. \dho{Despite it was introduced more than four decades ago} \cite{woodham-ps}, achieving high accuracy and computational efficiency still remains a challenge.

\begin{figure}
	\centering
	\includegraphics[width=1.0\linewidth]{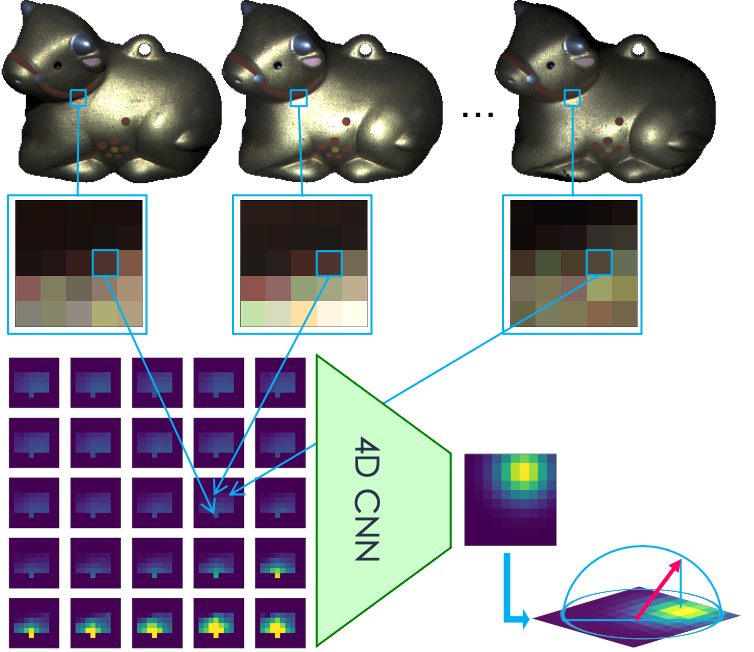}
	\caption{Our pipeline in a nutshell: A block of observation maps created from the local image patches around a pixel and known illumination directions are given as input to our 4D fully-convolutional network, which outputs a Gaussian heat-map. The surface normal can be estimated from this heat-map by a projection onto a unit hemisphere.}
	\label{fig:summary}
\end{figure}

Early model-based methods rely heavily on strong assumptions, such as convex surfaces, infinitely distant point light sources, orthogonal projection, linear response camera, and opaque materials whose reflectance properties can be described using an isotropic, constant or parametric bidirectional reflectance distribution function (BRDF). Over time, more elaborate techniques based on complex BRDFs were introduced to address poor performance caused by partial violations of these assumptions, such as cast shadows, inter-reflections, and ambient light sources. However, their relatively low representative power still falls short of accurately modeling real-world scenes, which limits their applicability.

Recent deep learning techniques allow to implicitly model more complex BRDFs and be more robust to the global illumination effects and other violations of the assumptions. However, the application of convolutional networks is not straightforward due to the potentially different number and position of lights used during training and inference phases.

The current state-of-the-art in calibrated photometric stereo adopts two distinct approaches to address this limitation. The first approach involves generating spatial features from each input image separately using a 2D convolutional network. This generates an a priori unknown number of feature activations, which are then fused with a pooling layer. Conversely, the second approach processes each spatial location on the object separately along the photometric (or illumination) axis. This is done by first transforming the individual pixel intensities into so-called \emph{observation maps} and then feeding these maps into a 2D convolutional regression network to estimate the 3D surface normals.

In contrast to these approaches that focus primarily on a single domain alone, we propose a new fully-convolutional and bias-free architecture that can simultaneously leverage both spatial and photometric context. As depicted in Figure~\ref{fig:summary}, our network architecture builds upon the observation maps but incorporates spatial information using both 2D and 4D separable convolutions to better capture global effects. Furthermore, we argue that instead of directly estimating the surface normals, posing the problem as a heat-map regression one both improves accuracy and results in almost an order of magnitude reduction in the number of parameters and multiply–accumulate operations (MACs) during inference. We show that our efficient architecture improves upon the state-of-the-art on the DiLiGenT benchmark and CyclesPS test datasets.





\begin{figure*}
    \centering
    \includegraphics[width=1.0\linewidth]{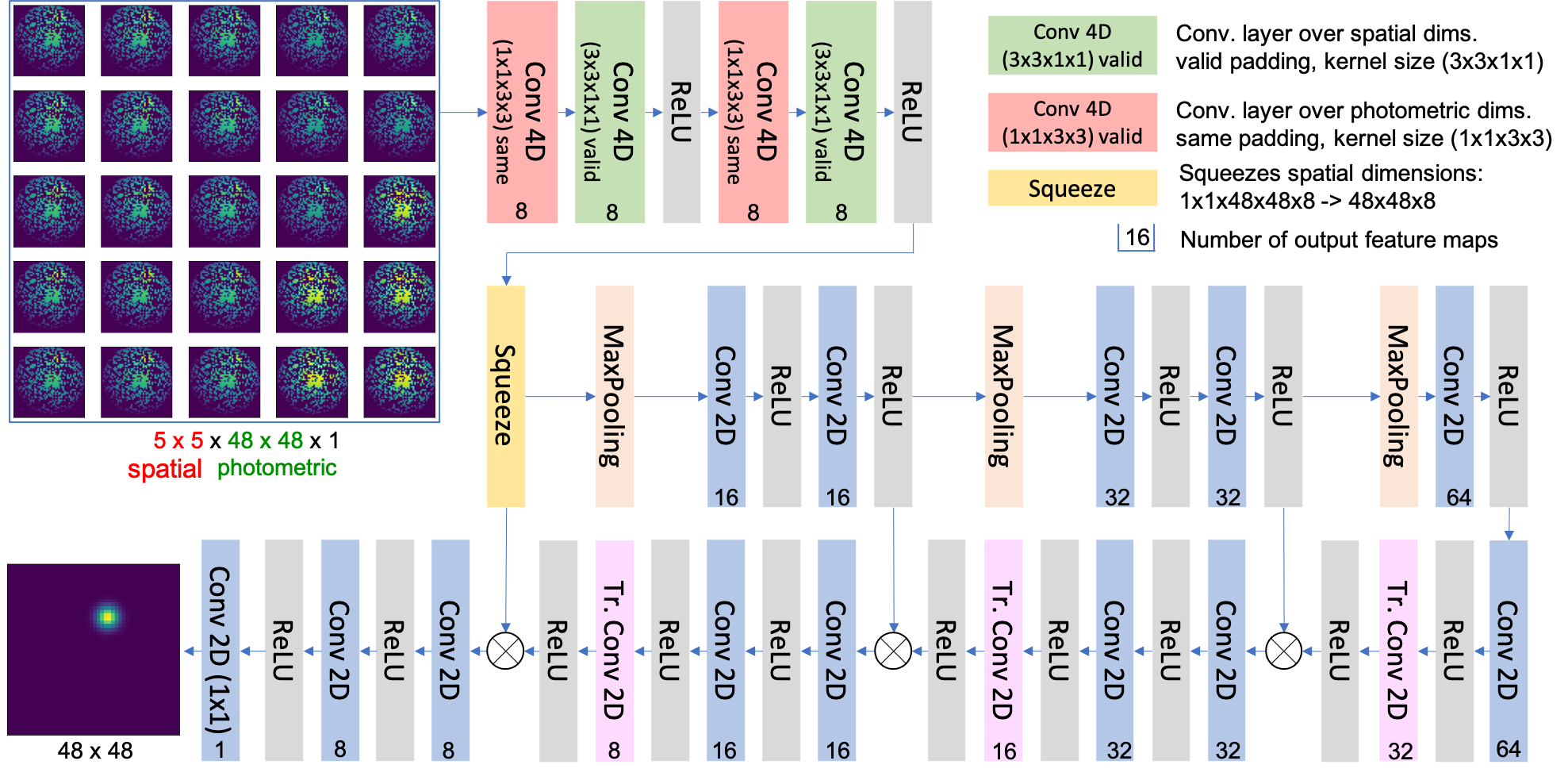}
    \caption{Our fully-convolutional and bias-free architecture processes 5-D observation maps comprised of both spatial and photometric dimensions. The output heat-maps encode the projection of the surface normals and are obtained with the same bijection used for creating the observation maps.}
    \label{fig:model}
\end{figure*}

\section{Related Work}
\label{sec:relatedwork}
Early model-based approaches in photometric stereo formulate the problem as point-wise inverse image rendering, which they solve for each spatial location separately. To make the problem tractable, they generally resort to strong assumptions such as convex surfaces and linear-response cameras with orthogonal projection. They also usually ignore chromaticity and rely on opaque isotropic BRDFs. As a result, in the special case of constant and diffuse BRDF the set of constraints leads to a basic image formation model:


\begin{equation}
    I_j = L_j \lambda max(n^T l_j, 0),    
\end{equation}
which describes the observed intensity $I_j$ as a function of the direction $l_j$ and intensity $L_j$ of the $j$-th light source, as well as the surface normal $n$ and a constant reflectance parameter $\lambda$. Given a sufficient number of observations of a surface point, the normals $n$ can then be estimated using the least-squares method as was first introduced by Robert J. Woodham \cite{woodham-ps} more than 40 years ago. However, the underlying assumptions are violated in practice for some of the observations.

More recent techniques address this limitation by removing outlier observations that can not be explained with their models \cite{mukaigawa2007analysis, wu2010robust, miyazaki2010median, wu2009photometric, ikehata2012robust}. However, their BRDF models still remain too simplistic to account for materials with complex reflectance properties. Using more advanced parametric BRDFs can alleviate this issue but leads to more complex optimization problems and still limits the application to a certain set of materials.  
\dho{Using a lookup table of example BRDFs \cite{discrete-search} or placing a reference object from the same material \cite{hertzmann2005example} into the scene addresses most of the BRDF-related issues, however, these methods are even more prone to errors caused by global illumination effects in complex scenes with non-convex surfaces.}

With the introduction of modern deep learning techniques in the field of photometric stereo, complex BRDFs could be implicitly modeled as latent features in neural networks. For instance, the pioneering work of Santo \etal \cite{dpsn} introduced the first fully-connected architecture for estimating surface normals from a set of intensity observations obtained with pre-defined light directions. Their approach involves  implicitly learning an inverse BRDF mapping from the observations to the normals, which is shown to be robust to cast shadows appearing on non-convex surfaces thanks to the dropout-like augmentations used during training. However, the method requires the number and direction of light sources to remain the same between training and inference, which severely limits its use.

Many recent CNN-based methods for calibrated photometric stereo utilize one of the two distinct approaches to deal with the a priori unknown number of input images and to incorporate the known illumination directions. The first approach involves first processing each input image separately along with its known illumination direction using a conventional 2D CNN architecture \cite{ps-fcn, ps-fcn-journal, attention-psn}. This produces a set of latent feature maps, which are subsequently fused using a pooling operation and then fed into a regression head network to generate per-pixel surface normal estimates. While this network architecture can successfully capture spatial cues among neighboring pixels, the pooling operation used to aggregate information from different illumination directions puts a strong restriction on the space of realistic BRDFs that can be estimated. 


The second approach tackles the problem of unknown illuminations by applying CNNs in the photometric domain \cite{cnn-ps, px-net, spline-net, learning-to-mitify-ps}. For each spatial location in the input images, it first projects the observed pixel values onto a 2D \emph{observation map} based on their respective illumination angles such that adjacent pixels in the observation map correspond to adjacent illumination angles. It then uses 2D CNNs to estimate a surface normal corresponding to the spatial location from the input observation map. The use of such an input that encodes essential photometric information in a compact way allows for better modeling of complex BRDFs and has been shown to improve performance. However, compared to the first approach, it lacks the spatial information, which can help in reasoning for global illumination effects such as shadows and inter-reflections.


In this paper, we advocate that bridging the gap between these two approaches by aggregating the information from the two input domains yields improved performance. We show that formulating the problem as a Gaussian heat-map regression one and using a fully-convolutional bias-free network architecture results in higher accuracy and more efficient inference.

\section{Method}
\label{sec:method}
In this section, we describe our bias-free fully-convolutional architecture for surface normal estimation using heat-maps. Our method is based on the concept of observation maps introduced in \cite{cnn-ps}. For each spatial position in the images, we create an observation map ${O \in \mathbb{R}^{w \times w}}$ from the individual observations $I_j$, illumination directions $l_j = [l^j_x, l^j_y, l^j_z] \in S^2$ (where $S^2$ is the space of 3\nobreakdash-dimensional vectors spanning a unit hemisphere), and light intensities $L_j$:
\begin{equation}
    O_{\text{int}(f(l_j))} = \frac{I_j}{L_j},
\end{equation}
where $\text{int}(\cdot)$ represents a rounding operation and f is a bijection $f: S^2 \rightarrow \mathbb{R}^2$:
\begin{equation}
\label{eq:bijection}
    f(l_j) = f([l^j_x, l^j_y, l^j_z]) = \left[w*\frac{l^j_x+1}{2}, w*\frac{l^j_y+1}{2}\right],
\end{equation}
Note that contrary to the original definition of the observation maps, we do not divide it by the maximal value as we use a bias-free, and therefore, scale equivariant network architecture \cite{mohan2019robust}. The bijection of \eqref{eq:bijection} allows mapping each sample consisting of multiple 2D images and their respective illumination directions to a 5D tensor ${\mathbf{I} \in \mathbb{R}^{\textit{height} \times \textit{width} \times w \times w \times c}}$ with the first two dimensions corresponding to the image space, the following two corresponding to the observation maps encoding photometric context, and the last representing the intensity spectrum. Similar to other works, we ignore chromaticity and set $c = 1$. Our approach can easily be extended to handle multi-channel inputs, however.

Processing such a 5D input with a fully convolutional architecture would allow capturing the entire spatial and photometric context, and would theoretically have the ability to model global illumination effects. However, it is impractical to process such a large data block with standard computer hardware, as even with an input image resolution of $512 \times 512$ and an observation map of size $48 \times 48$, the 5D tensor would take more than 2GB of memory.

We therefore propose an architecture that takes into account the entire photometric context, to properly model potentially complex BRDF, but only a local spatial context of size $b \times b$. Our fully-convolutional neural network depicted in Figure~\ref{fig:model} takes as an input a batch of patches $P \in \mathbb{R}^{b \times b \times w \times w \times c}$ and outputs a batch of two-dimensional heat-maps $M \in \mathbb{R}^{w \times w}$. We obtain the target heat-maps by applying the same orthogonal projection of \eqref{eq:bijection} to the normals $n$ for the spatial center of the $b \times b$ patch. To ease the optimization process, we train the network to produce a 2D Gaussian with standard deviation $\sigma=2$ centered at the projected location:
\begin{equation}
    M^{n}_{x,y} = \frac{\exp\left(-\frac{(x - f(n)_x)^2 + (y - f(n)_y)^2}{2\sigma^2}\right)}{2\pi\sigma}.
\end{equation}

At test time, the surface normal can be obtained by finding the coordinates of the maximum value $m_x, m_y$ and using the inverse projection \eqref{eq:bijection}:
\begin{equation}
    \label{eq:inversebijection}
    n = [n_x, n_y, n_z] = f^{-1}([m_x,m_y]).
\end{equation}
A straightforward approach to estimate the coordinates $m_x, m_y$ with sub-pixel accuracy is to fit a Gaussian to the predicted heat-maps and find its mean. However, for performance reasons, we estimate it as the center of the mass computed locally around the maximum coordinates.

To process the input 5D tensors, early layers of our network use separable 4D convolutions. However, since 4D convolutional layers are not yet efficiently implemented in existing deep learning frameworks, for performance reasons, we use two distinct 2D convolutional layers instead. The first one of these is applied over the photometric dimensions using \emph{same} padding and the second over the spatial ones using \emph{valid} padding since a only single output heat-map of size $w \times w$ is desired per sample. In this work, we use a spatial context of size $b=5$, and therefore we use only 2 separable convolutions to reduce these dimensions to $1 \times 1$. This is followed by a squeeze layer that outputs a 4D tensor and subsequently  standard 2D convolutional layers that operate in the photometric domain only.

Our architecture resembles a U-Net \cite{unet}, which comprises an encoder of $B$ convolutional blocks with max-pooling, and a decoder of the same number of blocks with transposed convolutions. The number of output feature maps of the convolutional layers is doubled with each down-scaling and halved with each up-scaling. To better propagate the gradients and preserve high-frequency information, the short-cut connections are added between each last layer of the encoder block and each first layer of the decoder block.
All the convolutional layers except the last one are followed by a ReLU activation function. In our experiments, we use $B=3$ blocks with $16$ output feature maps on the first level.

We omit the bias terms in all the layers of our network because this regularizes the network to prevent overfitting and makes it scale equivariant \cite{mohan2019robust}. This is useful especially in predicting the heat-maps, in which only the relative intensity differences play a role in estimating the surface normal directions.

\section{Experiments}
Obtaining accurate ground-truth data for real-world objects is a demanding task. Even the largest real photometric stereo dataset, DiLiGenT \cite{diligent}, contains only 10 objects. As these datasets are very scarce, they are used for evaluation purposes only. All existing supervised methods, therefore, rely on rendered training data and our method is no exception. In this section, we first describe our training and evaluation scheme and then present our results on two public datasets: DiLiGenT benchmark \cite{diligent} and CyclesPS test datasets \cite{cnn-ps}. We also provide an ablation study to demonstrate that the two main contributions of this work, namely the use of a joint spatio-photometric context and the Gaussian heat-map regression, improve the results independent of each other.

\subsection{CyclesPS Training Dataset}
\begin{figure}
    \centering
    \includegraphics[width=0.8\linewidth]{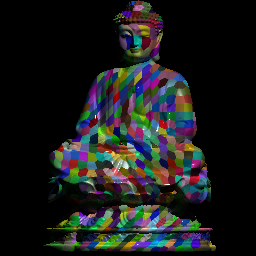}
    \caption{A sample image of size $256 \times 256$ from the training set of the CyclesPS \cite{cnn-ps} dataset.}
    \label{fig:cyclesps}
\end{figure}

We train our architecture on the training set of the CyclesPS dataset \cite{cnn-ps}. It consists of fifteen objects with complex geometry and three sets of Disney’s principled BRDFs involving metallic, specular, and diffuse materials. Each object with each set of BRDFs was rendered with roughly 1300 different directional lights (uniformly sampled from a hemisphere) using the \emph{Cycles} realistic renderer \cite{cycles}. To simulate the spatially varying BRDFs, which naturally appear in the wild, each scene was divided into about 5000 super-pixels where different BRDF characteristics were used as illustrated in Figure~\ref{fig:cyclesps}.

Note that such a simulation approach leads to very unrealistic objects in the spatial dimensions. While this doesn't pose a problem for entirely photometric approaches, it may be unfavorable for methods that rely on spatial context. Nevertheless, for small spatial contexts (e.g., 3 to 5 pixels), which we use in this work, this drawback is deemed negligible.

\subsection{Training Details}
Since the CyclesPS dataset is small the data augmentation is a vital step in training the neural network. This is  especially true for architectures that take a high dimensional input, and hence are more susceptible to overfitting. To this end, we adopt the data augmentation approach of the CNN\nobreakdash-PS method \cite{cnn-ps}. More specifically, we randomly keep 50 to 1000 illuminations for creating the observation maps, and we randomly threshold the maximal allowed angle between the camera and an illumination direction by $20^{\circ}$ to $90^{\circ}$ as the acute angles are usually more informative.

In this work, we assume an isotropic BRDF, which allows for encoding rotation equivariance into the network since rotating the input leads to rotated normals. We force it by enriching the training set with $K_\textit{train}$ rotations of the samples that are performed both spatially and photometrically. While for the spatial rotations, we use spline interpolation, the photometrical rotation can be achieved without any distortion by rotating the illumination directions before the creation of the observation maps. Similarly, we rotate the ground truth surface normal maps spatially, and we also rotate the individual normals.
In practice, we use $K_\textit{train} = 12$ rotations instead of 10 in the CNN-PS method \cite{cnn-ps}, as these can be implemented more efficiently with less distortion.

We train the network for 2 epochs on 90\% of the augmented training dataset and use the remaining samples for validation. We use the RMSprop optimizer with the initial learning rate of 0.001, to which we apply a decay of factor 0.985 every 1,000,000 samples. We use the mean squared error loss function on the heat-maps:
\begin{equation}
    \frac{1}{w^2B}\sum_{x,y,i}^{w,w,B}\left(M^{n^{pred}_i}_{x, y} - 100*M^{n^{gt}_i}_{x, y}\right)^2,
\end{equation}
where $B$ is the batch size. We developed the network in the TensorFlow Keras framework and trained it on 3 GPUs in parallel with a batch size of $B = 3\times256$, which took about 6 hours.

\begin{table*}[ht]
	\begin{center}
		\begin{tabular}{lcccccccccc|c}
			\toprule
			Method & Ball & Bear & Buddha & Cat & Cow & Goblet & Harvest & Pot 1 & Pot 2 & Reading & Average \\
			\midrule
			Baseline \cite{woodham-ps} & 4.10 & 8.39 & 14.92 & 8.41 & 25.60 & 18.50 & 30.62 & 8.89 & 16.65 & 19.80 & 15.39\\
			DPSN* \cite{dpsn} & \textbf{2.02} & 6.31 & 12.68 & 6.54 & 8.01 & 11.28 & 16.86 & 7.05 & 7.86 & 15.51 & 9.41 \\
			NIRGRPS$\ddagger$ \cite{neural-inverse-rendering} & 1.47 & 5.79 & 10.36 & 5.44 & 6.32 & 11.47 & 22.59 & 6.09 & 7.76 & 11.02 & 8.83 \\
			\midrule
			PS-FCN \cite{ps-fcn} & 2.82 & 7.55 & 7.91 & 6.16 & 7.33 & 8.60 & 15.85 & 7.13 & 7.25 & 13.33 & 8.39 \\
			Attention-PSN \cite{attention-psn} & 2.93 & 4.86 & 7.75 & 6.14 & 6.86 & 8.42 & 15.44 & 6.92 & 6.97 & 12.90 & 7.92 \\
			\midrule
			CNN-PS$\dagger$ \cite{cnn-ps} & 2.2  & 4.1  & 7.9  & 4.6  & 8.0  & 7.3  & 14.0 & 5.4   & 6.0  & 12.6 & 7.2 \\
			PX-NET$\dagger$ \cite{px-net} & 2.15 & 3.64 & 7.13 & 4.30 & 4.83 & 7.13 & 14.31 & 4.94 & 4.99 & 11.02 & 6.45 \\
			\midrule
			\textbf{Ours}$\dagger$ ($K_{\textit{test}}=10$) & 2.55 & 3.62 & 7.43 & 4.76 & 5.00 & 7.05 & 12.85 & 5.16 & 5.11 & 11.11 & 6.46 \\
			\textbf{Ours} ($K_{\textit{test}}=12$) & 2.49 & 3.59 & 7.23 & 4.69 & 4.89 & 6.89 & 12.79 & 5.10 & 4.98 & 11.08 & 6.37 \\
			\bottomrule
		\end{tabular}
	\end{center}
	\caption{The self-reported performance (mean angular error in degrees) of deep learning based methods on DiLiGenT main dataset. The baseline is the simple Lambertian approach. *DPSN network was trained for the specific light configuration of this dataset. \dho{$\ddagger$NIRGRPS is an unsupervised method that optimizes the normals at test time.} Methods marked with $\dagger$ use $K_{\textit{test}}=10$ rotations during the evaluation.}
	\label{tab:results-diligent}
\end{table*}

\subsection{Evaluation Details}
As we enforce the isotropy constraint during training, we can also benefit from it at inference time. Similar to other works \cite{cnn-ps, px-net}, we use $K_{\textit{test}}$ rotations of the inputs, inversely rotate the predicted normals, and average them to get the final surface normal estimate. In line with the training phase, we use $K_{\textit{test} = 12}$ in our experiments.

As described in \ref{sec:method}, our network predicts Gaussian heat-maps, from which we estimate the normals by finding the location with the maximal value \eqref{eq:inversebijection}. To achieve sub-pixel accuracy, we crop a $5 \times 5$ patch around this location in the heat-map (we reflect the values in case of borders) and compute a new sub-pixel accurate location as the center of the mass. Nevertheless, given the fact that we already average the estimated normals over $K_{\textit{test} = 12}$ rotations, the improvement brought by this step is a minor one.

We evaluate all methods in terms of the mean angular error (measured in degrees) between the ground truth and predicted normal directions.

\subsection{Evaluation on the DiLiGenT Benchmark Dataset}
\label{sec:diligent}

The \emph{DiLiGenT} dataset \cite{diligent} is the most commonly used real-world photometric stereo benchmark, which contains 10 scenes with various reflectance properties that spatially vary in some cases. Each scene is captured under 96 illumination directions which are known and differ slightly among the scenes. The resolution of the images is $612 \times 512$, but the objects span on average only 12\% of it. As reported in \cite{cnn-ps}, some objects in the dataset are problematic, which is why, similar to CNN-PS \cite{cnn-ps}, we ignore the first 20 images for the Bear object and flip the normals for the Harvest object.

We present the results of our method on the DiLiGenT dataset in Table~\ref{tab:results-diligent} and compare it to several leading techniques including CNN-PS \cite{cnn-ps} and PX-NET \cite{px-net}. Our method also significantly outperforms the CNN-PS \cite{cnn-ps} method that uses the same training dataset, but relies only on the photometric context. It also outperforms PS-FCN \cite{ps-fcn} and Attention-PSN \cite{attention-psn} methods that fuse spatial feature maps with max-pooling layers along the photometric dimension. Our results are on par with the ones of PX-NET \cite{px-net} for the baseline case of $K_\textit{test}=10$, and slightly better for $K_\textit{test}=12$. In terms of the number of parameters and MACs, however, our network is almost an order of magnitude smaller  than that of PX-NET as shown in Table~\ref{tab:params}.

It is worth noting that PX-NET relies on a practically infinite dataset as it randomly samples the normals, BRDFs, and light sources on the fly during training. However, at the time of writing this article, their dataset and rendering code were not publicly available to train on. While our approach is orthogonal to the PX-NET one, as one of the motivations for random sampling of normals was to remove the spatial correlation, we believe that online rendering of patches instead of pixels would improve the performance of our method even further and provide a fairer comparison.

\begin{figure*}
    \centering
    \includegraphics[width=1.0\linewidth]{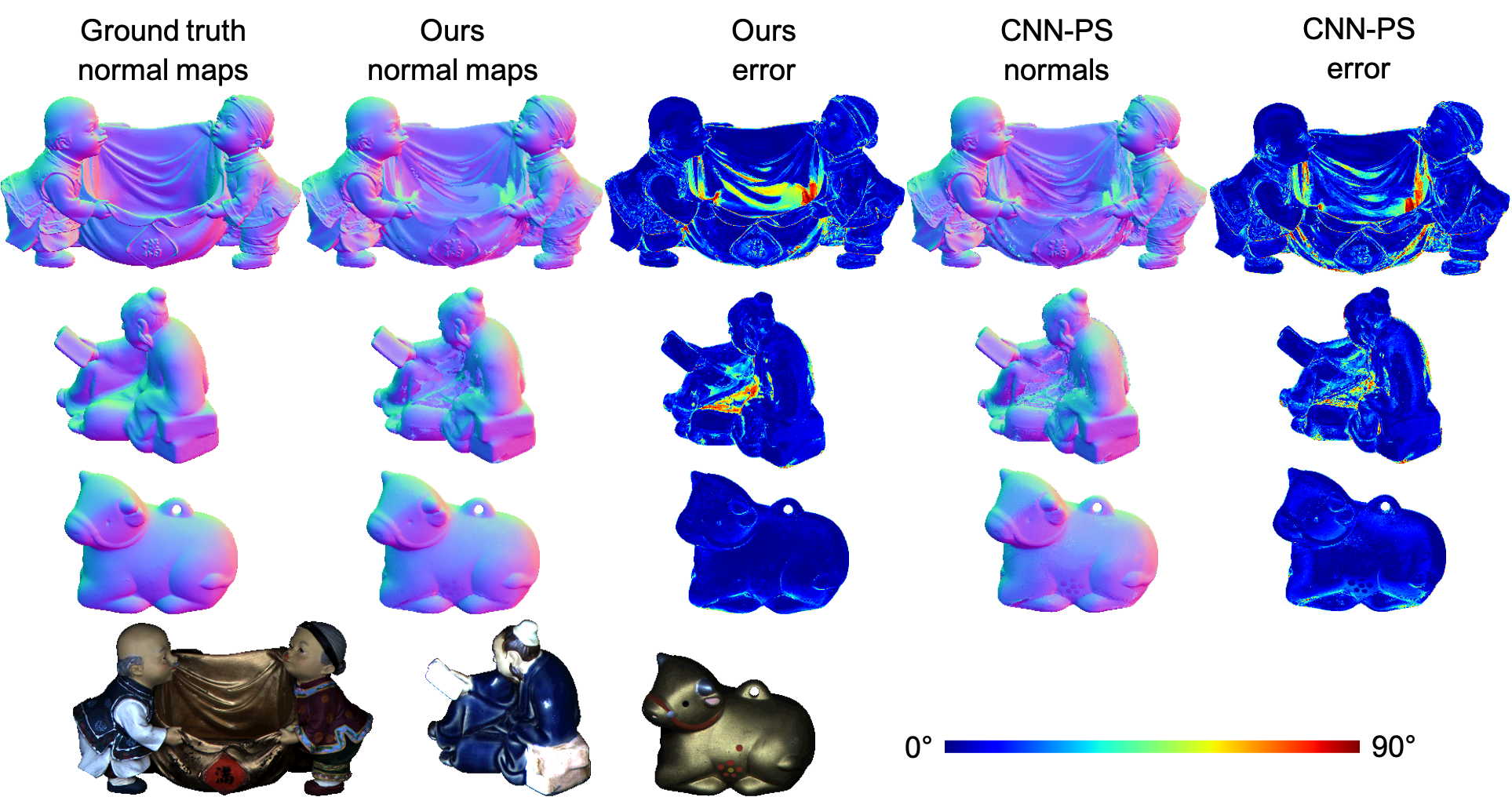}
    \caption{Ground truth and predicted normal maps for our and CNN-PS \cite{cnn-ps} methods and their respective angular errors in degrees. From top to bottom, the results for the Harvest, Reading, and Cow objects and their representative color pictures.}
    \label{fig:diligent-images}
\end{figure*}

A more detailed analysis of the results reveals that our method achieves significantly better performance over CNN-PS on complex shiny objects such as Cow, Harvest, or Reading. Figure~\ref{fig:diligent-images} illustrates the predictions for the three objects, which show that our predictions are smoother thanks to the incorporation of the spatial context. The most problematic regions for both approaches (and for the others evaluated in this work) are the large areas that are often entirely in shadow or have a lot of inter-reflections. While the small $5 \times 5$ spatial context used in this work is insufficient to accurately model such large surface areas, thin hinge lines are modeled fairly well with our method. 

We evaluate the impact of the number of rotations $K_{\textit{test}}$ used for inference by varying the $K_{\textit{test}}$ from $1$ to $20$ for both our method and CNN-PS. Table~\ref{tab:ktest} summarizes the results, which show that while our method greatly benefits from adding additional rotations at inference time, this effect is not as pronounced with the other methods. The accuracy of our method saturates around $K_{\textit{test}} = 12$. We argue that higher $K_{\textit{test}}$ might be affected by slight interpolation distortions and therefore result in slightly worse error values.

\begin{table}[ht]
    \begin{center}
        \begin{tabular}{lcc}
        \toprule
        Method & Parameters & MACs\\
        \midrule
        PX-NET \cite{px-net} & 4.5M & 9M \\
        CNN-PS \cite{cnn-ps} & 2.7M & 5M \\
        \midrule
        Ours & 0.5M & 1M \\
        \bottomrule
        \end{tabular}
    \end{center}
    \caption{The number of parameters and multiply–accumulate operations for three models evaluated.}
    \label{tab:params}
\end{table}

\begin{table}[ht]
    \begin{center}
        \begin{tabular}{lccc}
        \toprule
        $K_{\textit{test}}$ & Ours & CNN-PS & PX-NET \\
        \midrule
        1  & 7.19 & 7.6 & 6.69 \\
        8  & 6.53 & 7.16*  & - \\
        10 & 6.46 & 7.2 & 6.45 \\
        12 & 6.37 & 7.17* & - \\
        16 & 6.39 & 7.18* & - \\
        18 & 6.35 & 7.18* & - \\
        20 & 6.38 & 7.15* & - \\
        \bottomrule
        \end{tabular}
    \end{center}
    \caption{The impact of the number of rotations $K_{\textit{test}}$ used for inference on the DiLiGenT dataset. *CNN-PS results for $K_{\textit{test}} \notin \{1,10\}$ were produced from the model trained and provided by its authors.}
    \label{tab:ktest}
\end{table}

\subsection{Evaluation on the CyclesPS Test Dataset}
\begin{table*}[ht]
	\begin{center}
		\begin{tabular}{lccc|ccc|c}
			\toprule
			BRDF set & \multicolumn{3}{c}{Specular} & \multicolumn{3}{c}{Metallic} & Average\\
			Object & Sphere & Turtle & Paperbowl & Sphere & Turtle & Paperbowl & \\
			\midrule
			Baseline \cite{woodham-ps} & 4.6 & 12.3 & 29.4 & 35.4 & 39.7 & 39.3 & 26.8 \\
			CNN-PS \cite{cnn-ps} $K_{\textit{test}} = 10$ & 3.3 & 9.9  & 20.1 & 9.0 & 17.8 & 34.2 & 15.7 \\
			Ours $K_{\textit{test}} = 10$ & 3.0 & 9.8 & 18.5 & 5.0 & 14.4 & 31.2 & 13.7  \\
			Ours $K_{\textit{test}} = 12$ & 2.9 & 9.4 & 18.8 & 5.1 & 14.1 & 31.0 & 13.6  \\
			\bottomrule
		\end{tabular}
	\end{center}
	\caption{Results on the L17 subset of the CyclesPS test dataset.}
	\label{tab:results-cycles-m17}
\end{table*}
\begin{table*}[ht]
	\begin{center}
		\begin{tabular}{lccc|ccc|c}
			\toprule
			BRDF set & \multicolumn{3}{c}{Specular} & \multicolumn{3}{c}{Metallic} & Average \\
			Object & Sphere & Turtle & Paperbowl & Sphere & Turtle & Paperbowl &  \\
			\midrule
			Baseline \cite{woodham-ps} & 5.0 & 12.6 & 28.8 & 44.5 & 40.2 & 37.0 & 28.0 \\
			CNN-PS \cite{cnn-ps} $K_{\textit{test}} = 10$ & 0.9 & 3.3 & 6.0 & 1.4 & 5.7 & 9.5 & 4.5 \\
			Ours $K_{\textit{test}} = 10$ & 0.8 & 4.9 & 5.6 & 1.2 & 8.4 & 12.7 & 5.6 \\
			Ours $K_{\textit{test}} = 12$ & 0.8 & 4.3 & 5.5 & 1.1 & 7.9 & 12.7 & 5.4 \\
			\bottomrule
		\end{tabular}
	\end{center}
	\caption{Results on the L305 subset of the CyclesPS test dataset.}
	\label{tab:results-cycles-m305}
\end{table*}

The \emph{CyclesPS} test dataset \cite{cnn-ps} comprises three rendered non-convex objects, which are generated in the same manner as the CyclesPS training dataset, except that its super-pixels are much larger (there are only 100 super-pixels per scene) and only two sets of BRDFs are used: specular and metallic. The synthetic dataset has two subsets that we refer to hereafter as L17 and L305. The first one is rendered with 17 uniformly sampled illumination directions while the latter with 305 directions. For both subsets, the images have a resolution of $512 \times 512$. However, unlike the DiLiGenT dataset, the objects are spread onto more than half of the pixels.

Tables~\ref{tab:results-cycles-m17} and~\ref{tab:results-cycles-m305} respectively give the results of our method and CNN-PS for the L17 and L305 subsets. While our method outperforms CNN-PS for the L17 subset, it is slightly worse on the L305 subset. This can be explained by the observation that while the spatial context plays an important role when the set of illuminations are sparse, its importance diminishes with a large number of illuminations  as the photometric domain already encompasses the necessary information for the estimation of the normals. While our method slightly under-fits for a large number of illuminations, it generalizes better to real-world objects with more practical illumination configurations.


\begin{table*}[ht]
    \begin{center}
        \begin{tabular}{lccccccccccc|c}
        \toprule
        Method & & Ball & Bear & Buddha & Cat & Cow & Goblet & Harvest & Pot 1 & Pot 2 & Reading & Average \\
        \midrule
        \textbf{CNN-PS} & avg & 2.90 & 4.02 & 8.15 & 4.76 & 7.66 & 7.63 & 14.69 & 5.43 & 6.69 & 12.82 & 7.47 \\
        & std & 0.40 & 0.10 & 0.15 & 0.29 & 0.44 & 0.09 & 0.35  & 0.28 & 0.21 & 0.52  & 0.22 \\
        \midrule
        \textbf{Spatial} & avg & 2.75 & 3.99 & 7.39 & 4.85 & 7.38 & 7.08 & 13.91 & 5.33 & 5.91 & 11.32 & 6.99 \\
        & std & 0.47 & 0.20 & 0.21 & 0.42 & 0.20 & 0.06 & 0.14 & 0.44 & 0.10 & 0.27 & 0.19 \\
        \textbf{Heat-map} & avg & 2.98 & 3.72 & 7.69 & 5.01 & 5.83 & 7.66 & 13.46 & 5.51 & 5.64 & 11.95 & 6.95 \\
        & std & 0.42 & 0.04 & 0.16 & 0.15 & 0.24 & 0.14 & 0.09 & 0.11 & 0.15 & 0.39 & 0.14 \\
        \bottomrule
        \end{tabular}
    \end{center}
    \caption{Ablation study showing the effect of the heat-maps and spatial processing on the DiLiGenT main dataset \cite{diligent}. The values are reported as the mean angular errors in degrees. It shows both the average and standard deviation over 5 runs.}
    \label{tab:ablation}
\end{table*}

\subsection{Ablation Study}
In this section, we demonstrate the impact of each of the two contributions separately -- heat-map prediction with bias-free U-Nets and incorporation of the spatial context. 
To this end, we constructed a standard U-Net network that takes as an input the 2D observation maps of size $w = 48$ and outputs the 2D heat-maps as in the 4D case. We use $B=3$ blocks and set the output number of feature maps in the first block to 8. The resulting network has about 130K parameters, which is 20 times lower than CNN-PS \cite{cnn-ps} to which we compare it against.
We train the network for 4 epochs using the RMSprop optimizer and the same learning rate decay strategy used in our main experiments.

To evaluate the relative impact of the spatial context, we took the CNN-PS network and replaced the first layer with a 4D separable convolutional one introduced in Section~\ref{sec:method}. We kept the rest of the original CNN-PS network unchanged. Since we have only a single 4D convolution, we use a patch size of $b=3$. To be consistent with the original paper, we train the network using the Adam optimizer for 10 epochs. However, due to performance and memory constraints, we had to augment the training data by $K_{\textit{train}}=8$ rotations instead of the original $10$. The reason is that spatial rotations contrary to the photometric ones need interpolation which is computationally costly.

We denote the first and the second networks described above as \emph{Heat-map} and \emph{Spatial} respectively. As in the main experiments reported in Table~\ref{tab:results-diligent}, we train them on the CyclesPS train dataset and report the results on the DiLiGenT dataset. Table~\ref{tab:ablation} summarizes the results for the original CNN-PS approach as well as the Heat-map and Spatial ones. We have run each experiment 5 times with different random seeds, and we report both the average and the standard deviation over the runs. For all the methods, we used $K_{\textit{test}}=10$. 
The results show that both contributions lead to improved performance over the CNN-PS baseline. Note that the discrepancy between the reported errors in the CNN-PS \cite{cnn-ps} paper and the measured values in the table is likely caused by a customized training loop (e.g. performing the last iterations on convex, diffuse or specular, surfaces) that was not reported in the original paper.

\section{Conclusion}
We introduced a novel neural network architecture for the general calibrated photometric stereo problem. Our approach makes use of both the spatial and the photometric context by employing 4D convolutional layers and regresses to Gaussian heat-maps that encode the surface normal directions. Experimental results demonstrate that these two strategies are complementary to each other and they achieve state-of-the-art results on two public photometric stereo datasets when used together. Future research will focus on ways of overcoming the memory constraints and more holistic modeling of the spatial context to better account for the global illumination effects.
    
\section{Acknowledgments}
We thank Satoshi Ikehata for providing us with the training dataset and for clarifying the training loop of the CNN\nobreakdash-PS method. We also thank Siavash A. Bigdeli for all fruitful discussions.

{\small
\bibliographystyle{ieee_fullname}
\bibliography{psbib}
}

\end{document}